\documentclass[manuscript,screen]{acmart}

\AtBeginDocument{%
  \providecommand\BibTeX{{%
    \normalfont B\kern-0.5em{\scshape i\kern-0.25em b}\kern-0.8em\TeX}}}

\setcopyright{acmlicensed}
\copyrightyear{2018}
\acmYear{2018}
\acmDOI{XXXXXXX.XXXXXXX}

\acmJournal{JACM}
\acmVolume{37}
\acmNumber{4}
\acmArticle{111}
\acmMonth{8}

\acmISBN{978-1-4503-XXXX-X/18/06}

\usepackage[marginal]{footmisc}
\usepackage{tabularx}
\usepackage{multirow}
\usepackage{alphabeta}
\usepackage{amsthm}
\usepackage{amsmath}
\usepackage{booktabs} 
\usepackage{threeparttable}
\usepackage{bbding}
\usepackage{xpatch}
\usepackage{pifont}
\usepackage{wasysym}

\usepackage{amssymb}
\usepackage{svg}
\usepackage{makecell}
\usepackage{amsmath,amsfonts}
\usepackage{algorithmic}
\usepackage{algorithm}
\usepackage{array}
\newcolumntype{C}[1]{>{\centering}p{#1}}
\usepackage[caption=false,font=normalsize,labelfont=sf,textfont=sf]{subfig}
\usepackage{textcomp}
\usepackage{stfloats}
\usepackage{url}
\usepackage{verbatim}
\usepackage{graphicx}
\usepackage{soul}
\usepackage{bm}

\newcommand{\cmmnt}[1]{}
\begin{document}

\title{Machine Unlearning for Traditional Models and Large Language Models : A Short Survey}

\author{Yi Xu}
\email{yixu00@ruc.edu.cn}
\affiliation{%
  \institution{Renmin University of China}
  \streetaddress{Zhongguancun St. 59th}
  \city{Haidian}
  \state{Beijing}
  \country{China}
  \postcode{100872}
}

\renewcommand{\shortauthors}{Yi Xu}

\begin{abstract}
With the implementation of personal data privacy regulations, the field of machine learning (ML) faces the challenge of the "\textit{right to be forgotten}". Machine unlearning has emerged to address this issue, aiming to delete data and reduce its impact on models according to user requests. Despite the widespread interest in machine unlearning, comprehensive surveys on its latest advancements, especially in the field of Large Language Models (LLMs) is lacking. This survey aims to fill this gap by providing an in-depth exploration of machine unlearning, including the definition, classification and evaluation criteria, as well as challenges in different environments and their solutions. Specifically, this paper categorizes and investigates unlearning on both traditional models and LLMs, and proposes methods for evaluating the effectiveness and efficiency of unlearning, and standards for performance measurement. This paper reveals the limitations of current unlearning techniques and emphasizes the importance of a comprehensive unlearning evaluation to avoid arbitrary forgetting. This survey not only summarizes the key concepts of unlearning technology but also points out its prominent issues and feasible directions for future research, providing valuable guidance for scholars in the field.
\end{abstract}



\maketitle

\section{Introduction}
In the rapidly evolving digital era, the management and protection of personal data have become paramount concerns for individuals and organizations alike. The advent of comprehensive data privacy laws across the globe reflects a collective commitment to safeguarding the sensitive information that fuels today's data-driven economy, which can be found in The European Union’s General Data Protection Regulation (\textit{GDPR})~\cite{noauthor_general_2018}, the California Consumer Privacy Act (\textit{CCPA})~\cite{noauthor_california_2018}, the Act on the Protection of Personal Information (\textit{APPI})~\cite{webpage:APPI}, and Canada’s proposed Consumer Privacy Protection Act (\textit{CPPA})~\cite{webpage:CPPA}. Central to these legal frameworks is the enshrinement of the \textit{"right to be forgotten"}, empowering individuals to demand the erasure of their personal data from the repositories of companies and service providers.

This legal mandate poses a significant challenge to the field of machine learning (ML), where datasets are foundational to the development and refinement of predictive models \cite{ma_learn_2023, chen_evaluating_2021}. The necessity to honor the requests while preserving the integrity and utility of ML models has led to the emergence of a novel discipline: machine unlearning \cite{mercuri_introduction_2022}. At its core, machine unlearning focuses on the methodologies and algorithms that enable the selective removal \cite{guo_certified_2020} or de-emphasis of data from ML models \cite{chen_novel_2019}, ensuring compliance with privacy regulations without unduly compromising performance \cite{cao_towards_2015}.

Despite the burgeoning interest and rapid advancements in machine unlearning, the academic and industrial communities have yet to converge on a unified understanding of the field's scope, techniques, and applications. This has resulted in a fragmented landscape, with pockets of research progressing in relative isolation. To bridge this divide and foster a more cohesive approach to machine unlearning, there is an urgent need for a thorough and systematic survey that consolidates the current state of knowledge.

This survey aims to fill this gap by providing a structured and comprehensive overview of the field of machine unlearning. We delve into the intricacies of the taxonomy of unlearning algorithms, exploring both centralized and distributed frameworks. We engage in a critical examination of the concept of approximate unlearning, weighing the trade-offs between data privacy and model efficacy.

Furthermore, we assess the current evaluation criteria, scrutinizing their effectiveness in measuring the success of unlearning endeavors. Considering the prevalance of Large Language Models (LLMs)(\cite{zhao2023survey, openai2022chatgpt, achiam2023gpt}), people finds that LLMs are vulnerable to some questions which may leak privacy \cite{yao2024survey, ghayyur2023panel} or even pose threat to public welfares \cite{karamolegkou2023copyright}, so applying machine unlearning for LLMs is significantly vital for LLM training. Methods like red-teaming \cite{ge2023mart, yu2023gptfuzzer}, data augmentation \cite{yuan2023llm} and other protection of privacy \cite{jang2022knowledge, si2023knowledge} has been introduced, so we also analyse the machine unlearning methods applied to LLMs, serving as a complement to the existing surveys.

We also recognize the diverse applications of machine learning, identify and discuss the unique challenges and proposed solutions associated with unlearning in various contexts. This survey not only synthesizes the existing body of research but also charts a course for future investigation. We anticipate that this work will serve as a valuable reference for researchers, practitioners, and policymakers alike, as they navigate the intricate balance between harnessing the power of ML and upholding the sanctity of data privacy. By addressing the state-of-art LLM's machine unlearning, we hope this work can help researchers to dig deeper into the protection of both individual, business and even nation's data, privacy in the future AI development, avoiding the possible privacy catastrophe caused by LLMs or even the AGI.
\begin{figure}
    \centering
    \includegraphics[width=0.6\linewidth]{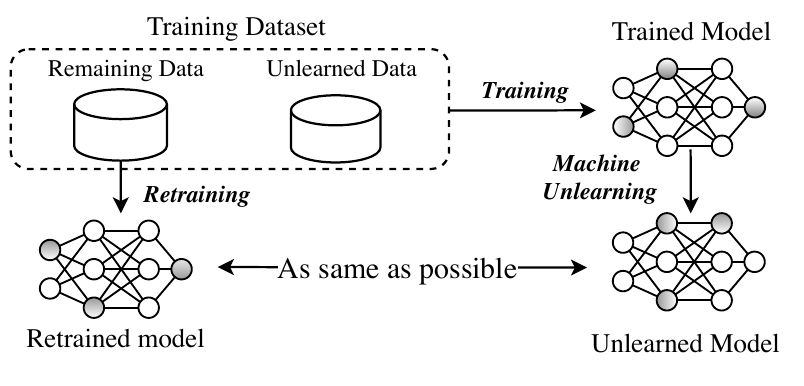}
    \caption{Illustration of Machine Unlearning. Depiction from survey \cite{xu_machine_2023}.}
    \label{fig:mu:example}
\end{figure}



\subsection{Motivations}
Machine unlearning, also known as selectively forgetting, data deletion, or scrubbing, necessitates the complete and swift removal of samples and their influence from both a training dataset and a trained model. This process isn't solely driven by regulatory requirements and legal mandates; it also arises from the privacy and security concerns of data providers, as well as the needs of model owners themselves. Indeed, eliminating the influence of outlier training samples from a model can enhance its performance and robustness significantly. There are several techniques to protect data privacy, like Differential Privacy \cite{zhou_privacy-preserving_2020, wei2021low, zhu2017differentially} , Data Masking \cite{xiang2021high}, Online Learning \cite{zhang2022counteracting, zhang2021counterfactual} and Catastrophic Forgetting \cite{liu2021overcoming, chen2021overcoming}, but all of them are different from machine unlearning to some extent \cite{xu_machine_2023}.
When users revoke permissions over certain training data, simply removing those data from the original training dataset is inadequate, as attackers can still extract user information from the trained models \cite{ullah_machine_2021}. One apparent solution to completely erase information from the model is to retrain it from scratch (depicted as the retraining process in Figure~\ref{fig:mu:example}). However, many complex models are constructed using vast sets of samples, rendering retraining a computationally intensive process \cite{cao_towards_2015, bourtoule_machine_2021}. Furthermore, in specific learning scenarios such as federated learning \cite{yang2019federated}, where the training dataset may not be directly accessible, retraining becomes unfeasible. Hence, to mitigate computational burdens and ensure the feasibility of machine unlearning across all scenarios, innovative techniques must be developed (illustrated as the unlearning process in Figure~\ref{fig:mu:example}).

\subsection{Contributions of this survey}
This survey provides a thorough view in both traditional and current LLM's machine unlearning. We make a substantial contribution by offering a comprehensive perspective on both traditional and contemporary approaches to machine unlearning within Large Language Models (LLMs). By exploring the evolution of machine unlearning techniques in LLMs, the survey provides valuable insights into the historical development and the latest advancements in this field. This holistic view enables researchers and practitioners to gain a deeper understanding of the challenges, opportunities, and emerging trends in LLM machine unlearning. Through its thorough examination of traditional methods alongside cutting-edge approaches, the survey equips stakeholders with the knowledge needed to navigate the complexities of machine unlearning in the context of LLMs effectively.

\subsection{Organization of the survey}

Our survey is organized as follows. Section 2 presents the preliminaries of machine unlearning (MU). Section 3 summarizes the taxonomy of MU, includes the discussion of the difference between traditional unlearning and LLM unlearning, and evaluation criteria, as well as targets of the MU. Section 4-7 discuss the data-driven MU, model-based MU, parameter-tuning LLM MU and parameter-agnostic LLM MU methods with the fine-grained categories and explanations respectively. In the last part, section 8 discuss both the challenges MU are currently facing, and the future of MU.

\section{Preliminaries}

\subsection{Definitions of the symbols}
Follow up the setting of \cite{xu_machine_2023, li_machine_2024}, we introduce the notations that will be consistently applied in this survey. We proceed to outline the typical lifecycle of a machine learning model to clarify the stages at which the unlearning process is implemented. Subsequently, we examine the various objectives of unlearning strategies and highlight the distinctions between machine unlearning and alternative approaches.

\begin{table}
  \caption{Notations.}
  \label{tab:Notations}
  \resizebox{\textwidth}{!}{
  \begin{tabular}{c|c||c|c}
    \toprule
    Notations &  Definition & Notations & Definition\\
    \midrule
     $\mathcal{D}$          &The original training dataset 	            &$\mathcal{D}_{r}$      &The remaining dataset \\
     $\mathcal{D}_{u}$      &The unlearning dataset             &$\bm x_{i}$	    & Single sample in $\mathcal{D}$ \\
     $\mathcal{X}$          &The sample space 	            &$\mathcal{Y}$ 	        &The label space \\
     $y_{i}$		        &The label of sample $\bm x_{i}$&$N$				    &The size of $\mathcal{D}$\\
     $\bm x_{i,j}$     &The $j$-th feature in $\bm x_{i}$   &$d$                    &The dimension of $\bm x_{i}$\\
     $\mathcal{M}$ & The learned model & $\mathcal{A}(\cdot)$   &The learning process \\
     $\mathcal{U}(\cdot)$   &The unlearning process & $\mathcal{R}(\cdot)$   &The retraining process \\
     $\theta$  &The parameters of learned model & $\theta_u$	&The parameters of unlearned model  \\
     $\theta_r$	 &The parameters of retrained model & $\sigma$ & The modified parameter of learned model \\
    \bottomrule	
	\end{tabular}
	}
\end{table}

\subsection{Definitions of the Machine Unlearning}

We defined the vector variables as bold lowercase, e.g., $\bm x_i$, the \textit{sample vector}, and spaces or sets as italicized uppercase, e.g., $\mathcal{D}$, the \textit{dataset}. Definition of machine learning is provided based on a supervised learning setting. The instance space is defined as $\mathcal{X} \subseteq \mathbb{R}^{d}$, with the label space defined as $\mathcal{Y} \subseteq \mathbb{R}$. $\mathcal{D}=\left\{\left(\bm x_{i}, y_{i}\right)\right\}_{i=1}^{N}$ represents the original training dataset, where each sample $\bm x_{i} \in \mathcal{X}$ is a $d$-dimensional vector, and corresponding label $y_{i} \in \mathcal{Y}$, and $N$ is the size of $\mathcal{D}$. Let $d$ be the dimension of $\bm x_{i}$ and let $\bm x_{i,j}$ denote the $j$-th feature in the sample $\bm x_{i}$.

Now we provide the definition of machine unlearning.
\begin{definition}
    \label{def:mu}
    Consider a set of samples, namely the \textit{unlearning dataset}, i.e., $\mathcal{D}_{u}$, that we intend to eliminate from both the original training data and the resulting trained model, i.e., $\mathcal{D}_u = \mathcal{D} - \mathcal{D}_r$, where $\mathcal{D}_r$ is the \textit{retained dataset} and $\mathcal{D}$ is the \textit{original training dataset}. The unlearning process, $\mathcal{U}(\mathcal{A}(\mathcal{D}), \mathcal{D}_u, \mathcal{D})$, is characterized as a function that accepts a model $\mathcal{M}$ trained on a dataset $\mathcal{D}$, i.e., $\mathcal{M} \leftarrow \mathcal{A}(\mathcal{D})$ and an unlearning dataset  $\mathcal{D}_{u}$, and outputs the unlearned model parameter series $\mathbf{\theta}_u$. Note that we can simply retrain the model, i.e., $\mathcal{R}(\mathcal{D}_r)$ to get the retrain parameter series $\theta_{r}$. The final goal is to keep the unlearn model's parameter series and performance as similar as the retrained model's, i.e. $\mathcal{U}(\mathcal{A}(\mathcal{D}), \mathcal{D}_u, \mathcal{D}) \approx \mathcal{R}(\mathcal{D}_r)$. This revised model $\mathbf{\theta}_{u}$ is crafted to perform in such a way that it behaves as though it has no knowledge of the unlearning dataset $\mathcal{D}_{u}$.
\end{definition}

\section{Taxonomy, evaluation metrics and targets of the machine unlearning}

Table~\ref{tab:taxo:summary} provides a comprehensive overview of the general taxonomy of machine unlearning and its evaluation criteria, highlighting the targets examined in this survey. The taxonomy is meticulously crafted, drawing inspiration from the intricate design details of various unlearning strategies. Approaches that focus on altering the training dataset are neatly classified under the umbrella of data-driven machine unlearning. In contrast, techniques that involve the direct adjustment of the weights within a trained model are aptly designated as model-based machine unlearning. Since we have to handle with the state-of-the-art LLM machine unlearning, which quite differs from traditional ones, we categorize it into parameter-tuning \& agnostic LLM unlearning. For the purposes of evaluation, our initial categorization distinguishes between empirical and theoretical methodologies. In the last part, targets of machine unlearning are introduced.

\begin{table}[]
\caption{Summary of taxonomy, evaluation metrics and targets of the machine unlearning}
\label{tab:taxo:summary}
\scalebox{0.8}{
\renewcommand{\arraystretch}{2}
\begin{tabular}{|cccc|}
\hline
\multicolumn{1}{|c|}{\multirow{6}{*}{Traditional Machine Unlearning}} & \multicolumn{1}{c|}{\multirow{3}{*}{Data-driven machine unlearning}} & \multicolumn{2}{c|}{Data influence/poisoning}                                    \\ \cline{3-4} 
\multicolumn{1}{|c|}{}                                                & \multicolumn{1}{c|}{}                                                & \multicolumn{2}{c|}{Data partition}                                              \\ \cline{3-4} 
\multicolumn{1}{|c|}{}                                                & \multicolumn{1}{c|}{}                                                & \multicolumn{2}{c|}{Data augmentation}                                           \\ \cline{2-4} 
\multicolumn{1}{|c|}{}                                                & \multicolumn{1}{c|}{\multirow{3}{*}{Model-based machine unlearning}} & \multicolumn{1}{c|}{\multirow{2}{*}{Model-agnostic methods}} & Model shifting    \\ \cline{4-4} 
\multicolumn{1}{|c|}{}                                                & \multicolumn{1}{c|}{}                                                & \multicolumn{1}{c|}{}                                        & Model pruning \\ \cline{3-4} 
\multicolumn{1}{|c|}{}                                                & \multicolumn{1}{c|}{}                                                & \multicolumn{1}{c|}{Model-intrinsic methods}                 & Model replacement     \\ \hline
\multicolumn{1}{|c|}{\multirow{3}{*}{LLM Machine Unlearning}}         & \multicolumn{1}{c|}{\multirow{2}{*}{Parameter-tuning LLM unlearning}} & \multicolumn{2}{c|}{Parameter optimization}                                      \\ \cline{3-4} 
\multicolumn{1}{|c|}{}                                                & \multicolumn{1}{c|}{}                                                & \multicolumn{2}{c|}{Parameter merging}                                           \\ \cline{2-4} 
\multicolumn{1}{|c|}{}                                                & \multicolumn{1}{c|}{Parameter-agnostic LLM unlearning}               & \multicolumn{2}{c|}{In-context unlearning (ICuL)}                                    \\ \hline
\multicolumn{1}{|c|}{\multirow{5}{*}{Evaluation Criteria}}            & \multicolumn{1}{c|}{\multirow{4}{*}{Empirical Evaluation}}           & \multicolumn{2}{c|}{Time-based}                                                  \\ \cline{3-4} 
\multicolumn{1}{|c|}{}                                                & \multicolumn{1}{c|}{}                                                & \multicolumn{2}{c|}{Accuracy-based}                                              \\ \cline{3-4} 
\multicolumn{1}{|c|}{}                                                & \multicolumn{1}{c|}{}                                                & \multicolumn{2}{c|}{Similarity-based}                                            \\ \cline{3-4} 
\multicolumn{1}{|c|}{}                                                & \multicolumn{1}{c|}{}                                                & \multicolumn{2}{c|}{Attack-based}                                                \\ \cline{2-4} 
\multicolumn{1}{|c|}{}                                                & \multicolumn{1}{c|}{Theoretical Evaluation}                          & \multicolumn{2}{c|}{Theory-based}                                                \\ \hline
\multicolumn{1}{|c|}{\multirow{2}{*}{Targets of Machine Unlearning}}  & \multicolumn{3}{c|}{Exact Machine Unlearning}                                                                                                           \\ \cline{2-4} 
\multicolumn{1}{|c|}{}                                                & \multicolumn{3}{c|}{Approximate Machine Unlearning}                                                                                                     \\ \hline
\end{tabular}

}
\end{table}

\subsection{Traditional unlearning taxonomy}
In this part, we discuss two main categories of traditional unlearning taxonomy: \textit{Data-driven machine unlearning, Model-based machine unlearning}.

\subsubsection{Data-driven machine unlearning}
Data-driven machine unlearning is a sophisticated technique employed by model providers to facilitate the unlearning of data through the strategic restructuring of the training dataset. This approach encompasses a trio of distinct processing methods, each tailored to different modes of reorganizing the data: \textit{influence/poisoning}, \textit{partition}, and \textit{augmentation}, as detailed in the works like \cite{bourtoule_machine_2021} and \cite{graves2021amnesiac}.
\begin{itemize}
    \item Data influence/poisoning:
    Model providers manipulate the training data of a machine learning model in a way that alters the model's behavior or predictions. For instance, when one of the class should be removed, providers can offer wrong samples of this class to allow the model forget class's features, thus achieve the unlearning goal. \cite{graves2021amnesiac} is the most famous example in this type.
    \item Data partition:
    In data pruning, the model provider initiates the process by partitioning the training dataset into multiple distinct sub-datasets. Upon these sub-datasets, a series of sub-models are independently trained. Once receiving an unlearning request, the model provider proceeds to purge the targeted samples from the respective sub-datasets that include them and subsequently retrains the impacted sub-models. The elegance of this approach lies in its sub-models, where the prunes of unlearning are confined to the affected sub-datasets post-segmentation, rather than retrain the whole model using the whole dataset. This strategic segmentation ensures that the unlearning process is localized, which can be applied to distributed machine learning methods or federated learning, thereby maintaining the integrity and performance of the overall model framework. The most famous example of data partition method is the \textit{SISA} \cite{bourtoule_machine_2021}, and several derivations \cite{chen_recommendation_2022, yan_arcane_2022, fan_fast_2023},  has been introduced.
    \item Data augmentation:
    In data augmentation, the model provider intentionally substitutes the training dataset $\mathcal{D}$ with a newly augmented dataset, denoted as $\mathcal{D}_{\text{aug}} \leftarrow \mathcal{D}$. This augmented dataset $\mathcal{D}_{\text{aug}}$ is subsequently employed to train a model, facilitating the implementation of unlearning upon receiving an unlearning request. For instance, \cite{shan2020fawkes} introduces the Fawkes system, which aims to protect personal privacy against unauthorized deep learning models through data augmentation techniques. Fawkes achieves privacy protection by adding subtle, adversarial perturbations to the original data, which are virtually invisible to the human eye but sufficient to mislead deep learning models. The method emphasizes the importance of data augmentation in enhancing data privacy protection, especially in the use and sharing of image data.
\end{itemize}

\subsubsection{Model-based machine unlearning} In model-based machine unlearning, model manipulator aims to achieve the unlearn goal by modifying model's parameters or even deleting/adding some of the layer of the model. Main types can be categorized into \textit{Model-agnostic} and \textit{Model-intrinsic}. When it comes to detailed methods, we give the \textit{shifting, pruning} and \textit{replacement} \cite{xu_machine_2023}.
\begin{itemize}
    \item Model shifting:
    In model shifting, model manipulator modify the model's internal parameters to counteract the effects of the data that needs to be forgotten, denote as $\theta_u = \theta + \sigma$, where $\theta$ represents the original parameters of the model, and $\theta_u$ is the expected unlearned parameters, $\sigma$ is the added modification parameters or values. The essence of model shifting is to figure out how each piece of data affects the model's settings and then make changes to eliminate that influence. However, this is no easy task. For complicated models, like \textit{deep neural networks} (DNNs), it's really hard to pinpoint exactly how a single piece of data affects the overall recipe. Because of this, many methods that use model shifting make certain assumptions (includeing \textit{influence unlearning} \cite{koh2017understanding, guo_certified_2020}, \textit{fisher unlearning} \cite{martens_new_2020, golatkar_eternal_2020}) to get the unlearning job done.  For example, \cite{wu_deltagrad_2020} introduced the \textit{DeltaGrad} method, applying gradient modifiers to make the model shifting. These methods can be applied to LLM machine unlearning, since the prevalence of LLM training is fine-tuning or RLHF based methods \cite{liu2024towards, yao2023large}.

    \item Model pruning:
    In model pruning, model manipulators use the pruning techniques to remove some parameters that have connection with the data to forget, denote as $\theta_{u}\leftarrow\theta/\sigma$, where $\theta_{u}$ is the unlearned parameters, $\theta$ is the original parameters, and $\sigma$ is the parameters which should be removed. This types of unlearning often shows up in federated unlearning, like the \textit{FedEraser} \cite{liu_federaser_2021} and \textit{class-discriminative pruning} \cite{wang_federated_2022}.
    
    \item Model replacement:
    In model replacement methods, model manipulator save some pre-calculated parameters, denote as $\theta_{pre}$, and aggregate the static parameters, denote as $\theta_{static}$, as the final unlearned parameters, that is $\theta_{u}\leftarrow\theta_{pre}\cup\theta_{static}$. These methods are usually depends on the specific type of the models, such as decision tree or random forest tree, like \cite{schelter_hedgecut_2021}.

\end{itemize}

\subsection{LLM unlearning taxonomy}
Following \cite{si_knowledge_2023}, we discuss two main categories of LLM unlearnining taxonomy: \textit{Parameter-tuning unlearning}, \textit{Parameter-agnostic unlearning}.
\subsubsection{Parameter-tuning LLM unlearning}
\begin{itemize}
    \item Parameter optimization:
    The most direct unlearning approach adopted by LLM providers is to emulate traditional model-based unlearning in machine unlearning - parameter optimization. In parameter optimization methods, various proposed approaches aim to better serve the specific parameters of the model to selectively modify the LLM's generative and output behaviors, while avoiding damage to the LLM's performance on other normal tasks. The latest work \cite{eldan_whos_2023, yao2023large, maini2024tofu, liu2024towards} are usually directly related to supervised fine-tuning (SFT) and reinforcement learning with human feedback (RLHF). Some earlier works \cite{wang2023kga, jang2022knowledge} are simpler, directly using reverse gradients \cite{franceschi2017forward} for updates, which enables LLM to gradient ascend through the data-to-forget, thus achieving the goal of unlearning. In addition, there are some works that, by imitating the previous model replacement approach, obtain redundant parameters, which can respond quickly when specific unlearning requirements arise, thereby eliminating direct modifications to existing effective parameters and preventing negative impacts on the performance of other tasks. 
    \item Parameter merging:
    LLM providers often also need more economical unlearning methods, so many works are also focusing on the method of parameter merging. This method is simpler than parameter optimization; it usually only involves simple arithmetic operations on parameters (usually addition and subtraction), and there is no need for online updates, thus saving a large number of parameter training steps and costs. If there is a need to fine-tune models that have been deployed, this method is the most practical and effective. However, parameter merging still requires a certain computational cost, and the effectiveness of simply overlaying model parameters is still debatable, with more theoretical analysis yet to be introduced. For example, \cite{ilharco2022editing} proposes a new paradigm for steering the behavior of neural networks, centered around \textit{task vectors}. A \textit{task vector} specifies a direction in the weight space of a pre-trained model, such that movement in that direction improves performance on the task, and thus improves the unlearning procedure.
    
\end{itemize}

\subsubsection{Parameter-agnostic LLM unlearning}
\begin{itemize}
    \item In-context unlearning (ICuL):
    As the most popular method currently, In-context Learning (ICL) also has a similar approach in unlearning: In-context Unlearning Learning (ICuL) \cite{pawelczyk2023context}. The biggest feature of this method is that it no longer focuses on direct modifications to parameters, that is, a simplest parameter-agnostic unlearning learning method. Specifically, ICuL treats the LLM as a black-box model and combines ICL's context learning ability to forget the data that needs to be forgotten. The advantage of ICuL is that it no longer requires adjusting the model's parameters, the cost is very low, and it provides the possibility of unlearning the LLM anytime, anywhere. However, the drawback is that the adjustment effect may be limited to a single context conversation, and the model still contains sensitive or even harmful knowledge content that should be forgotten.
\end{itemize}

\subsection{Evaluation criteria}
When it comes to evaluation criteria, existing surveys often separate into 2 general types, including empirical and theoretic evaluation. Our survey follows the former techniques and update them in LLM machine unlearning methods. Table~\ref{tab:eval} shows the advantages and drawbacks of different evaluation metrics.

\subsubsection{Time-based Metrics}
\begin{itemize}
    \item Unlearn time:
    The amount of time required for an unlearning request.
    \item Relearn time:
    Number of epochs needed for the unlearned model to achieve the accuracy of the source model.
\end{itemize}

\subsubsection{Accuracy-based Metrics}

Accuracy of the unlearned model on both the unlearning set $\mathcal{D}_u$ and remaining set $\mathcal{D}_r$.
\subsubsection{Similarity-based Metrics}

\begin{itemize}
    \item Completeness:
    Overlapping (e.g., Jaccard distance) of output space between the retrained and unlearned models.
    \item Layer-wise Distance
    Weight difference between the original model and the retrain model.
    \item Activation Distance :
    Average L2-distance between the predicted probabilities of the unlearned model and the retrained model on the unlearning set.
    \item $JS$-Divergence:
    Jensen-Shannon divergence between the predictions of the unlearned and retrained models.
    \item Epistemic Uncertainty \cite{hullermeier2021aleatoric}: Quantifies the amount of information the model exposes.
\end{itemize}

\subsubsection{Attack-based Metrics}

\begin{itemize}
    \item Membership Inference Attack \cite{khosravy2022model}: Recall: $ \frac{\texttt{\#detected items}}{\texttt{\#forget items}} $.
    \item $ZRF$ Score \cite{chundawat_can_2023}:
    $ZRF = 1 - \frac{1}{nf}\sum\limits_{i=0}^{n_f} {JS}(M(\bm x_i), T_d(\bm x_i))$, measures the intentionality of the unlearned model's outputs on the forget items.
    \item Model Inversion Attack \cite{shokri2017membership}: Utilizes visualization for qualitative verifications and evaluations.
\end{itemize}

\subsubsection{Theory-based Metrics}

Anamnesis Index ($AIN$) \cite{chundawat_zero-shot_2023}: $AIN = \frac{r_t (M_u, M_{orig}, \alpha)}{r_t (M_s, M_{orig}, \alpha)}$, used in \textit{zero-shot machine unlearning }\cite{chundawat_zero-shot_2023} to evaluate model performance with theoretical analysis.

\begin{table}[!htbp]
\caption{Summary of the Evaluation criteria.}
\label{tab:eval}
\scalebox{0.55}{
\renewcommand{\arraystretch}{1.5}
\begin{tabular}{|ccl|l|l|}
\hline
\multicolumn{3}{|l|}{\textbf{Evaluation criteria}}                                                                                                  & \textbf{Advantages}                                                               & \textbf{Drawbacks}                                                                                                                           \\ \hline
\multicolumn{1}{|c|}{\multirow{12}{*}{Empirical Evaluation}} & \multicolumn{1}{c|}{\multirow{2}{*}{Time-based}}       & Unlearn Time                & Easy to judge the effectiveness of the unlearn procedure                          & Need comparison with the retrained model                                                                                                     \\ \cline{3-5} 
\multicolumn{1}{|c|}{}                                       & \multicolumn{1}{c|}{}                                  & Relearn Time                & Easy to understand                                                                & Other metrics cannot be guaranteed                                                                                                           \\ \cline{2-5} 
\multicolumn{1}{|c|}{}                                       & \multicolumn{1}{l|}{\multirow{2}{*}{Accuracy-based}}   & Accuracy on $\mathcal{D}_u$           & Easy to implement                                                                 & \begin{tabular}[c]{@{}l@{}}Whether model have remove the sensitive data\\  or not is unknown\end{tabular}                                    \\ \cline{3-5} 
\multicolumn{1}{|c|}{}                                       & \multicolumn{1}{l|}{}                                  & Accuracy on $\mathcal{D}_r$            & Easy to implement                                                                 & \begin{tabular}[c]{@{}l@{}}Result on remained data is not convincing \\ without the result on unlearned data\end{tabular}                    \\ \cline{2-5} 
\multicolumn{1}{|c|}{}                                       & \multicolumn{1}{c|}{\multirow{5}{*}{Similarity-based}} & Completeness                & Easy to confirm that the model has forget the data                                & Need accuracy to confirm the model doesn't shift to much                                                                                     \\ \cline{3-5} 
\multicolumn{1}{|c|}{}                                       & \multicolumn{1}{c|}{}                                  & Layer-wise Distance         & Easy to confirm the layer's parameter changes                                     & \begin{tabular}[c]{@{}l@{}}Low absolute error of layer-wise parameter \\ doesn't guarantee the similar action made by the model\end{tabular} \\ \cline{3-5} 
\multicolumn{1}{|c|}{}                                       & \multicolumn{1}{c|}{}                                  & Activation Distance         & Similar with layer-wise distance                                                  & Similar with layer-wise distance                                                                                                             \\ \cline{3-5} 
\multicolumn{1}{|c|}{}                                       & \multicolumn{1}{c|}{}                                  & $JS$-Divergence               & Easy to confirm the generated results' contribution                               & Need accuracy to confirm the model doesn't shift to much                                                                                     \\ \cline{3-5} 
\multicolumn{1}{|c|}{}                                       & \multicolumn{1}{c|}{}                                  & Epistemic Uncertainty       & Can check the unlearned model's uncertainty                                       & Need accuracy to confirm the model doesn't shift to much                                                                                     \\ \cline{2-5} 
\multicolumn{1}{|c|}{}                                       & \multicolumn{1}{c|}{\multirow{3}{*}{Attack-based}}     & Membership Inference Attack & Conclusion can be obtained intuitively                                            & Damage model's performance                                                                                                                   \\ \cline{3-5} 
\multicolumn{1}{|c|}{}                                       & \multicolumn{1}{c|}{}                                  & $ZRF$ Score                   & \begin{tabular}[c]{@{}l@{}}Avoid unlearned model just give wrong or random output \\ when given forgotten items\end{tabular} & \begin{tabular}[c]{@{}l@{}}Controversy to some design of unlearning algorithms, \\ especially in LLM unlearning\end{tabular}                 \\ \cline{3-5} 
\multicolumn{1}{|c|}{}                                       & \multicolumn{1}{c|}{}                                  & Model Inversion Attack      & Conclusion can be obtained intuitively                                            & Implementation is difficult                                                                                                                  \\ \hline
\multicolumn{1}{|c|}{Theoretical Evaluation}                 & \multicolumn{1}{c|}{Theory-based}                      & $AIN$ Score                   & \begin{tabular}[c]{@{}l@{}}Can check both the uncertainty and efficiency \\ at the same time\end{tabular}  & Only effective on zero-shot machine unlearning                                                                                               \\ \hline
\end{tabular}

}
\end{table}

\subsection{Targets of the machine unlearning}

Training a model from the ground up using $\mathcal{D}_r$ is simplistic and frequently unfeasible in real-world situations because of its significant computational and time requirements. In centralized machine learning environments, contemporary machine unlearning algorithms strive to tackle the challenge of resource intensiveness while guaranteeing the removal of $\mathcal{D}_u$. These algorithms are divided into two main categories depending on whether they utilize a retraining process: exact unlearning and approximate unlearning. A comparison between these two categories is illustrated in Table \ref{tab:goal:comparison}.

\begin{table}[!htbp]
\caption{Goals and Comparison of Different Targets.}
\label{tab:goal:comparison}
\resizebox{\textwidth}{!}{
\begin{tabular}{|l|l|l|l|}

\hline
\textbf{Target} & \textbf{Goal}  & \textbf{Advantages} & \textbf{Drawbacks}\\ 
\hline
Exact unlearning  & \begin{tabular}[c]{@{}l@{}}Indistinguishable distribution between retrained\\ and unlearned model\end{tabular}           & \begin{tabular}[c]{@{}l@{}}Keeps the best accordance to the original model and the retrained model,\\ Hard for attackers to recover sensitive data\end{tabular} & \begin{tabular}[c]{@{}l@{}}Hard to design, \\ Takes a lot of tine to train, \\ Hard to implements to multi scenes\end{tabular}\\ 
\hline
Approximate unlearning & \begin{tabular}[c]{@{}l@{}}Generated data contains no sensitive data, \\ Do not require the unlearned model's distribution to \\ keep as same as the retrained model\end{tabular} & \begin{tabular}[c]{@{}l@{}}Easier to train and test,\\ Can be applied to multi scenes with lower cost by using unified backbones\end{tabular} & \begin{tabular}[c]{@{}l@{}}Cannot guarantee the removal of sensitive parameters,  \\ Privacy data may still leave in the model\end{tabular} \\ 
\hline
\end{tabular}
}
\end{table}

\section{Data-driven Machine unlearning}
In this section, we discuss how data-driven machine unlearning are recognized, and gives the earlier to latest works that use data-driven methods to unlearn. We separate data-driven methods into three parts: \textit{data influence/poisoning}, \textit{data partition}, and \textit{data augmentation}.

\subsection{Data influence/poisoning}
In most model attacks or information theft, many attackers assume that the model has overfitted to the training data during training, and therefore, they can use incorrect data to mislead the model, obtain a modified output data distribution, and trick the model into bypassing testing and directly outputting the original input content to guess the likely true input data \cite{hu2021membership, shokri2017membership}. Thus, the most direct approach is to consider the risk of data influence/poisoning at the beginning of training and include it as an objective of model goal after training. This method is simple but efficient, and many early works have used this approach, since it just need the model providers to manipulate the unlearn dataset $\mathcal{D}_u$.

\cite{graves2021amnesiac} proposed \textit{amnesiac unlearning}, which  put forward a framework for machine unlearning that involves randomly altering the labels of sensitive instances and subsequently retraining the model on this altered dataset through multiple iterations to achieve unlearning. ~\cite{felps_class_2021} have also adopted a similar strategy by deliberately poisoning the labels within the unlearning dataset and conducting fine-tuning on this corrupted dataset. Nevertheless, these unlearning techniques merely disrupt the association between the model's predictions and the instances, and do not completely eliminate the information about each sample that may still be encoded in the model's parameters.

\cite{tarun_fast_2023} have delineated the unlearning procedure into a two-stage approach, termed as \textit{impair} and \textit{repair}. During the initial stage, they train an error-maximization noise matrix to identify and focus on samples that are most impactful for the class targeted for unlearning. This noise matrix acts in contrast to the unlearning data, effectively obliterating the information associated with the data to be unlearned, enabling the model to forget one or multiple classes. To counteract any decline in model performance resulting from the unlearning process, the subsequent \textit{repair} phase involves further fine-tuning the model using the data that remains after the unlearning step. \cite{chundawat_zero-shot_2023} then extend this \textit{repair} and \text{impair} procedure into the zero-shot setting, which enables the model to unlearn with no samples. 

Furthermore, \cite{chen_boundary_2023} proposed the \textit{boundary unlearning}, which develop two novel boundary shift methods, namely Boundary Shrink and Boundary Expanding, both of which can rapidly achieve the utility and privacy guarantees. Key idea of \textit{boundary unlearning} is to shift the decision boundary of the original DNN model to imitate the decision behavior of the model retrained from scratch.

\subsection{Data partition}
The data partition method is based on the premise that when a portion of the data needs to be forgotten from the originally fully trained model, the entire model must be rebuilt from scratch. However, by leveraging the concept of ensemble learning \cite{dong2020survey}, where multiple sub-models are combined to form the final primary model for output, it becomes possible to retrain only the specific sub-models that utilize the data to be forgotten when the need arises. By ensuring that the performance of these sub-models remains consistent or close to the original, the performance of the primary model can remain virtually unchanged. This training approach is also highly suitable for distributed machine learning \cite{verbraeken2020survey} and even federated unlearning \cite{yang2019federated, wu_federated_2022}, which is why there has been significant attention given to this class of unlearning methods.

\cite{bourtoule_machine_2021} introduced a \textit{`sharded, isolated, sliced and aggregated'} framework (\textit{SISA} for short), which simply apply the conception of partitioning training data for sub-models. In addition to \textit{SISA}, \cite{chen_recommendation_2022} introduced a improved partition method, which uses attention-based aggregation to avoid the quick degradation of model performance when sub-model increases, and apply it to recommender systems, confirming the validity in practical use.

Moreover, \cite{yan_arcane_2022} proposed the \textit{ARCANE} unlearning architecture, based on the idea similar to \textit{SISA}. \textit{ARCANE} leverages ensemble learning to transform the retraining process into multiple one-class classification tasks, and also proposes data preprocessing techniques to further minimize retraining overhead and accelerate unlearning, including representative data selection, training state saving, and sorting to handle varying unlearning request distributions.
\subsection{Data augmentation}
Data augmentation-based unlearning algorithms primarily focus on using augmented data to enable the model to remember additional data distribution characteristics, thus quickly adapting to unlearning demands and preventing attackers from discerning distribution features. This approach is similar to data influence/poisoning, but the difference lies in the timing and nature of the data involved. Influence/poisoning occurs after the model has been trained and the data is often harmful or confusing, whereas data augmentation provides additional feature representations on the original dataset, with an emphasis on helping the model understand the data itself and adapt to unlearning with efficiency and effectiveness.
\cite{cao_towards_2015} presented a general, efficient unlearning approach by transforming learning algorithms used by a system into a summation form. To forget a training data sample, our approach simply updates a small number of summations -- asymptotically faster than retraining from scratch. It can also be applied to all stages of machine learning, including feature selection and modeling. In addition to this work, \cite{shibata2021learning} proposed a novel approach to machine learning that addresses the challenge of updating models when certain data needs to be forgotten or when new data must be incorporated without retraining the entire model from scratch. The proposed method allows for selective forgetting by identifying and adjusting the parameters associated with the data that needs to be forgotten, while retaining the knowledge relevant to the remaining data. This selective forgetting process is designed to be efficient and effective, ensuring that the model's performance is not compromised and that it can adapt to new information quickly. The article presents a method that balances the retention of valuable knowledge with the ability to forget information that is no longer relevant or desired, which is crucial for applications where data privacy and model adaptability are key concerns.

\subsection{Summary of Data-driven Machine unlearning}

Data-driven machine unlearning methods, such as data influence/poisoning, data partition, and data augmentation, each offer unique approaches to managing the forgetting process in machine learning models. Data influence/poisoning involves introducing new data to counteract or overwrite the influence of the data to be forgotten after the original training, which can be efficient but carries the risk of embedding harmful data. Data partition techniques, on the other hand, segment the dataset and retrain only the necessary parts, saving computational resources but requiring careful data management to avoid data overlap. Data augmentation enriches the training set with modified versions of existing data, improving the model's robustness and ability to generalize, yet it may not directly target the forgetting of specific data points. Each method presents trade-offs between computational efficiency, directness of forgetting, and the potential impact on the model's overall performance, and their effectiveness depends on the scale and complexity of the data, as well as the specific requirements of the application.

\section{Model-based Machine Unlearning}
During the model training phase, the objective is to establish a model that accurately reflects the relationship between the training dataset's inputs and the model's corresponding outputs. Consequently, directly altering the model to eliminate certain relationships can be an effective approach for unlearning specific data. Since currently we mainly focus on LLM unlearning, which are derived from model-based machine unlearning method, we only give the summary and discuss LLM unlearning in the following part.

\subsection{Summary of Model-based Machine Unlearning}
In essence, model-shifting based unlearning methods strive for enhanced efficiency through assumptions about the training process, including the datasets and optimization strategies employed. These methods, effective for simpler models like linear regression, face increased complexity with complex deep neural networks. Model pruning, a part of the unlearning process, involves significant architectural changes that could impact the performance of the models post-unlearning. It's important to mention that model replacement techniques necessitate pre-calculation and storage of all potential parameters for swift parameter replacement during unlearning. Therefore, developing more efficient unlearning strategies that balance model utility, storage requirements, and the effectiveness of the unlearning process are critical areas for further research.

\section{Parameter-tuning Large Language Model Unlearning}

\subsection{Parameter optimization}
Most works of the parameter optimization are related to gradient-based methods, e.g. gradient ascent, reversed gradient and reversed loss.
\cite{eldan_whos_2023} first introduced the LLM parameter optimization setting, by giving the instance of `Who's Harry Potter'. Specifically, it enhanced a model by further training it on specific target data to pinpoint the tokens most relevant to the unlearn goal, using a baseline model for comparison. In the following part, it generalized the target data by swapping unique phrases with more common ones and use the model's predictions to create new labels for each token, simulating what a model would predict. Finally, by fine-tuning the model on these new labels, it effectively removed the original text from the model's memory, ensuring it does not recall it when confronted with related context.

\cite{yu2023unlearning} introduced a novel method known as Partitioned Contrastive Gradient Unlearning (PCGU), a gray-box strategy designed for removing biases from pre-trained masked language models. PCGU focused on refining the weights that are primarily responsible for a particular type of bias by utilizing a first-order approximation derived from the gradients of contrasting sentence pairs. While the application of PCGU had been limited to the gender-profession domain, authors observed that it also contributed to a reduction in biases in other related areas. In addition to this, \cite{gu_second-order_2024} introduced a second-order optimization method to improve the performance. Authors presentd a fresh examination of the unlearning issue through the lens of second-order information, specifically the Hessian, as opposed to methods relying on first-order information. The unlearning techniques, drawing from the traditional Newton update, were designed to be independent of the data and model used, and had been demonstrated to effectively maintain utility and ensure privacy protection.

\cite{wang2023kga} introduced a general forgetting framework called KGA, aimed at inducing model forgetfulness. Unlike previous approaches that attempt to recover gradients or force models to perform close to one specific distribution, KGA maintains distribution differences (i.e., knowledge gap), relaxing the distribution assumption. Furthermore, we first apply the unlearning method to various NLP tasks (i.e., classification, translation, response generation) and propose several unlearning evaluation metrics with relevance. 

\cite{yao2023large} presentd a task of acheiving generalization, utility and efficiency in LLM unlearning by introducing an optimization goal which consider the forget, mismatch and maintain metrics of LLM, and then use gradient ascent method to update. Furthermore, \cite{maini2024tofu} proposed a new task of fictious unlearning (TOFU) and offered a dataset of 200 diverse synthetic author profiles, each consisting of 20 question-answer pairs, and a subset of these profiles called the forget set that serves as the target for unlearning, and used fine-tuning, gradient ascent with KL-divergence requirement to make LLM unlearn well on diverse datasets. Metrics of different benchmarks on TOFU are still frustrating, which shows that parameter optimization methods still needs improvement.
\subsection{Parameter merging}
Some works focus on tuning LLM for unlearning with a more compatible cost of training and testing, main techniques related to this type are applying arithmetic operations or simply remove/mask some layers/neurons to acheive the unlearning goal.
\cite{ilharco2022editing} proposed a novel approach to directing the actions of neural networks through the concept of task vectors, which define a trajectory in a model's weight space to enhance task performance. Task vectors were created by the difference in weights before and after task-specific fine-tuning. The study demonstrated that these vectors can be arithmetically manipulated, such as through negation and addition, to alter the model's behavior accordingly. Inverting a task vector reduced performance on the target task while maintaining performance on control tasks, and combining vectors can enhance performance across multiple tasks. Intriguingly, task vectors can also be merged based on analogical relationships to improve performance on a related task without direct training on that task.
\cite{zhang2024composing} proposed a method which involved combining parameter-efficient modules using linear arithmetic operations to integrate various capabilities without additional training. This approach uses addition and negation operators for flexible module composition and is applied for distribution generalization, multi-tasking, detoxifying, and domain transfer, with a successful case study on the Alpaca-LoRA model, enabling more efficient fine-tune unlearning on LLMs.
\cite{pochinkov2024dissecting}  introduced a selective pruning method for LLMs that removes neurons based on their relative importance on a targeted capability compared to overall network performance. This approach is a compute- and data-efficient method for identifying and removing neurons that enable specific behaviours. Authors also found that both feed-forward and attention neurons in LLMs are specialized in different tasks.

\subsection{Summary of Parameter-tuning Large Language Model Unlearning}

Parameter-tuning in Large Language Model (LLM) unlearning consists various techniques aimed at modifying model parameters to achieve forgetting. Overall, these methods show promise in achieving efficient and effective unlearning in LLMs, though there is still room for improvement, particularly in explicit evaluation metrics and the specialization of neurons for specific tasks and the interpretability of parameter-tuning.

\section{Parameter-agnostic Large Language Model Unlearning}
\subsection{In-context unlearning (ICuL)}
As the most special unlearning strategy, In-context unlearning (ICuL)\cite{pawelczyk2023context} didn't even touch the parameter of the LLM. To be specific, ICuL is designed for scenarios where only the model's API is accessible without access to its internal parameters. ICuL utilizes unlearning samples, such as input prompts, and their associated responses as prompt examples, to improve inference accuracy. By leveraging a constrained set of examples, ICuL aims to refine outputs through prompt-based learning.
\subsection{Summary of Parameter-agnostic Large Language Model Unlearning}

This method offers several distinct advantages. Firstly, it eliminates the need for fine-tuning the model's parameters, significantly reducing computational costs. Additionally, its low cost makes it accessible for widespread adoption, offering the flexibility to unlearn the Language Model (LLM) at any time and from any location. However, it's essential to note some limitations. While this method may effectively adjust the model within a specific contextual conversation, its impact could be confined to that particular context. Moreover, despite its unlearning capability, the model might still retain sensitive or potentially harmful knowledge, underscoring the importance of continuous refinement and oversight in AI development.

\section{Challenges and the future of machine unlearning}
Given the current research and discussions in the field of machine unlearning, future challenges lie in developing more efficient and accurate unlearning algorithms that can effectively erase or modify undesired memories without compromising the overall performance of the model. Moreover, as LLMs continue to advance, future research needs to focus on designing forgetting mechanisms that are resilient against malicious attacks and ensure the privacy protection throughout the unlearning process. Additionally, researchers should explore ways to quantify and assess the effectiveness of unlearning, establishing standardized evaluation metrics that can reliably measure and compare the efficacy of different unlearning techniques in practice, especially in newer area of unlearning, e.g. unbounded unlearning \cite{kurmanji2024towards}, more effective, safer LLM unlearning\cite{liu2024towards}. Future studies should also consider the legal and ethical implications \cite{usman2018exploring} of unlearning operations, ensuring that technological advancements align with societal values and regulatory requirements. Through these efforts, machine unlearning technology will better serve society by safeguarding individual privacy while promoting the responsible and sustainable development of artificial intelligence.

Expanding on this, the future of machine unlearning will likely involve interdisciplinary collaboration between AI researchers, legal experts, and ethicists to create a robust framework for forgetting that adheres to data protection laws and maintains the trust of users. There is a need for innovative solutions that not only address the technical aspects of unlearning but also consider the broader societal implications, such as the potential for misuse of power or the impact on freedom of information \cite{pozen2016freedom}. As AI systems become more integrated into daily life, the ability to forget information that is no longer necessary or relevant will become increasingly important. Therefore, investing in the research and development of machine unlearning is crucial for building AI systems that are adaptable, trustworthy, and respectful of individual rights and societal norms. The future of machine unlearning will be shaped by our ability to navigate these complex issues and create technologies that empower while protecting the interests of all stakeholders involved.

\section{Conclusion}
This survey addresses the critical challenge of the forgetting sensitive/poisonous data in the machine learning domain by delving into the advancements of machine unlearning, particularly within the context of Large Language Models (LLMs). By defining unlearning procedures, classifying unlearning approaches and establishing evaluation criteria, this work contributes significantly to the understanding and development of effective unlearning techniques in both traditional models and LLMs. This survey highlights the limitations of existing evaluation methods and underscores the necessity for comprehensive assessments to ensure the long-term effectiveness of unlearning, offering a solid foundation and clear directions for future research in this vital area of privacy protection.

\bibliographystyle{ACM-Reference-Format}
\bibliography{related_papers}

\end{document}